\documentclass[twoside,11pt]{article}
\usepackage[accepted]{melba}
\usepackage{color}

\usepackage{amsmath,amsfonts}

\definecolor{normal}{rgb}{.0,.0,.0}
\definecolor{myred}{rgb}{.8,.0,.0}
\definecolor{myblue}{rgb}{.0,.0,.8}

\newcommand{\autoa}{AutoA}
\newcommand{\autob}{AutoB}
\newcommand{\autoc}{AutoC}
\newcommand{\crowda}{CrowdA}
\newcommand{\crowdb}{CrowdB}
\newcommand{\crowdc}{CrowdC}
\newcommand{\expa}{ExpA}
\newcommand{\expb}{ExpB}
\newcommand{\expc}{ExpC}
\newcommand{\studa}{StudA}
\newcommand{\studb}{StudB}
\newcommand{\studc}{StudC}

\melbaheading{2021:020}{https://www.melba-journal.org/papers/2021-020.html}{2021}{1-26}{07/2021}{12/2021}{Raumanns, Schouten, Joosten, Pluim and Cheplygina}{}{}

\ShortHeadings{ENHANCE (ENriching Health data by ANnotations of Crowd and Experts)}{Raumanns et al.}
\firstpageno{1}

\title{ENHANCE (ENriching Health data by ANnotations of Crowd and Experts): \\ A case study for skin lesion classification}

\author{\name Ralf Raumanns \email ralf.raumanns@fontys.nl \\  
	\addr Fontys University of Applied Science, Eindhoven, The Netherlands \\
	\addr Eindhoven University of Technology, Eindhoven, The Netherlands
	\AND
	\name Gerard Schouten \email g.schouten@fontys.nl \\
	\addr Fontys University of Applied Science, Eindhoven, The Netherlands \\
	\addr Eindhoven University of Technology, Eindhoven, The Netherlands
	\AND
	\name Max Joosten \email m.p.h.joosten@student.tue.nl \\
	\addr Eindhoven University of Technology, Eindhoven, The Netherlands
	\AND
	\name Josien P. W. Pluim \email j.pluim@tue.nl \\
	\addr Eindhoven University of Technology, Eindhoven, The Netherlands
	\AND Veronika Cheplygina \email vech@itu.dk \\
	\addr IT University of Copenhagen, Denmark
}

\begin{document}

\maketitle

\begin{abstract}
We present \textbf{ENHANCE}, an open dataset with multiple annotations to complement the existing ISIC and PH2 skin lesion classification datasets. This dataset contains annotations of visual ABC (asymmetry, border, color) features from non-expert annotation sources: undergraduate students, crowd workers from Amazon MTurk and classic image processing algorithms. In this paper we first analyze the correlations between the annotations and the diagnostic label of the lesion, as well as study the agreement between different annotation sources. Overall we find weak correlations of non-expert annotations with the diagnostic label, and low agreement between different annotation sources. Next we study multi-task learning (MTL) with the annotations as additional labels, and show that non-expert annotations   improve the diagnostic performance of (ensembles of) state-of-the-art convolutional neural networks. We hope that our dataset can be used in and inspires further research into multiple annotations and/or MTL.

All data and models are available on: \url{https://github.com/raumannsr/ENHANCE}.

\end{abstract}


\section{Introduction}
Machine learning offers many opportunities, but medical imaging datasets, for example for skin lesion diagnosis, are limited, and overfitting can occur. To illustrate, Winkler and colleagues found that superimposed scale bars \citep{Winkler2021-cm} or skin markings \citep{Winkler2019-kn} in dermoscopic images may impair the diagnostic performance of the convolutional neural network (CNN) when unintentionally overfitting these artifacts during model training.

A promising approach to generalize better in small sample size settings is multi-task learning (MTL), where the model has to learn different tasks simultaneously. This approach showed improved performance in various medical applications, for example, for breast lesions \citep{Shi2019-xt,Liu2018-jy}. However, when moving from single-task to multi-task models, we need additional annotations. Applying MTL is challenging because datasets typically do not have such additional annotations. Furthermore, building a medical image dataset from scratch with expert annotations is time-consuming and costly.

We present a dataset of additional annotations for skin lesion diagnosis based on non-expert annotations on three dermatoscopic criteria: asymmetry, border and color (so-called ABC criteria). In dermatology, the use of the ABCDE (asymmetry, border, color, diameter, and evolution or elevation) rule is widespread. However, scoring the diameter (D) and evolution or elevation (E) are more complex tasks and therefore less suitable for non-expert annotation. The term \textit{non-expert} is defined here as annotations provided by three different annotation sources: undergraduate students, crowd workers from Amazon MTurk and automated annotations through classic image processing algorithms.

We study the quality of non-expert annotations from different viewpoints. Firstly, we determine the discriminative power of ABC features for diagnosis. We show to what extent the ABC annotations correlate to the diagnosis, and we study how we can use ABC annotations to improve the performance of a CNN. Secondly, the inter agreement level for A, B and C feature between the different annotation sources.
The study extends our research on the topic \citep{Raumanns2020-ac} by using automated annotations as well as comparing the performance on three open source CNN architectures, in particular: VGG-16 \citep{simonyan2015deep}, Inception v3 \citep{Szegedy2016-me} and ResNet50 \citep{He2016-ow} encoders. Further, we investigate whether MTL is also beneficial for automated annotations and show that the performance benefits from using multiple annotations in MTL.

Besides addressing the lack of expert annotations using non-expert ones, we make the dataset with collected ABC annotations and code open, eliminating obstacles for future research. More specifically, the investigation addresses the following research questions:
\begin{enumerate}
  \item \textbf{What is the correlation between the ABC annotations and the diagnostic label?} 
  \item \textbf{How can we use the ABC annotations to improve the performance of a CNN?}
  \item \textbf{What is the inter agreement level for A, B and C feature between the different annotation sources?}
  \item \textbf{How can CNN performance benefit from using multiple annotations?}
\end{enumerate}

Our results give valuable insights into the quality of non-expert annotations. Using the collected non-expert annotations in three different CNNs, we show that these are of added value for the performance of the models. This suggests that the use of non-expert annotations might be promising for application in similar domains. 

\section{Related Work}
The shortage of publicly available medical datasets suitable for machine learning is a widely shared problem. Less suitable datasets are small and contain few attributes. Several initiatives have been launched to disseminate and make more medical datasets accessible to alleviate this shortcoming. For example, \cite{Tschandl2018-sz} released the HAM10000 (``Human Against Machine with 10000 training images'') dataset containing dermatoscopic images from different populations. \cite{Pacheco2020-vz} released a skin lesion benchmark composed of clinical images collected from smartphone devices and a set of patient clinical data containing up to 21 features.

In machine learning, we usually concentrate on training a single-task model, focusing on a single output. These single-task models have been successful in medical imaging, but there is a potential to make them even more generalizable, thereby preventing overfitting by using MTL that share representations between related tasks. More specific MTL uses inductive transfer with domain-specific information to improve model generalization \citep{Caruana1997-hl}.

MTL has been successfully in various applications in medical imaging. \cite{Hussein2019-cq} presented a framework for the malignancy determination of lung nodules based on an MTL method using features provided by experts (radiologists). \cite{dhungel2017deep} used automatically extracted features in a two-step training process for mass classification in mammograms using a CNN and Random Forest model. \cite{murthy2017center} included shape information in classifying glioma nuclei and showed that they could improve CNN performance compared to a baseline model. Recently we \citep{Raumanns2020-ac} found that MTL ensembles with the VGG-16 encoder, combined with crowdsourced features, lead to improved generalization for skin lesion classification. For more background information on MTL in deep learning, we refer the reader to an overview written by \cite{ruder2017overview}.

In this work, we further investigate the use of non-expert annotations for skin lesion classification by comparing non-expert annotations and using them as additional output in three different multi-task models based on commonly employed CNN architectures. Note that the MTL approach allows using ABC annotations on the output side of the network, different from traditional approaches such as \cite{Giotis2013-iz,Cheplygina2018-ee}, where they are used as classifier inputs. 

\section{Dataset creation} 

This section describes the two public image datasets that we used and details the annotation procedures used to collect additional annotations for these datasets.

\subsection{Image datasets}
We used two publicly available datasets of dermoscopic images, the training set of the ISIC 2017 challenge dataset \citep{Codella2017-rd} and the PH2 dataset \citep{Mendonca2015-uc}. A summary of the both datasets is provided in Table \ref{tab:summary_datasets}. There are no other meta-data available than the characteristics presented.

\subsubsection{ISIC dataset}
The ISIC 2017 challenge was to develop lesion segmentation and lesion classification algorithms. The training dataset contains 2000 dermoscopic images with lesions of three classes: 374 melanoma, 1372 nevi, and 254 seborrheic keratosis lesions. Next to the diagnostic labels, the dataset also contains age and sex for some of the patients, binary lesion segmentation masks, and superpixel segmentation masks (the result of dividing a lesion image with the help of the SLIC algorithm \citep{Achanta2012-zk}).

\subsubsection{PH2 dataset} The PH2 dataset was acquired at the Dermatology Service of Hospital Pedro Hispano, Matosinhos, Portugal.  The objective was to create an open dataset to help comparative studies on dermoscopic images' segmentation and classification algorithms. The PH2 dataset contains 200 dermoscopic images, of which 80 common nevi, 80 atypical nevi, and 40 melanoma lesions. The dataset also contains information related to histology and diagnosis, and expert annotations of lesion characteristics including color, asymmetry, and pigment network.

\begingroup
\setlength{\tabcolsep}{10pt}
\begin{table}[ht]
    \centering
    \caption{Summary of the lesion characteristics in the ISIC and PH2 datasets. A dash (-) means that the information was not provided.}
    \label{tab:summary_datasets}
    \begin{tabular}{l l rr}
    \hline
    Characteristics & & ISIC & PH2 \\ [0.5ex]
    \hline
    Diagnosis & Benign nevi & 1372 & 160 \\
    & Melanoma & 374 & 40 \\
    & Seborrheic keratosis & 254 & 0 \\

    \hline
    Patient sex & Female & 871 & - \\
    & Male & 900 & - \\
    & Not provided & 229 & 200 \\
    Age group & 0-10 & 4 & - \\
    & 10-20 & 147 & - \\
    & 20-30 & 136 & - \\
    & 30-40 & 225 & - \\
    & 40-50 & 305 & - \\
    & 50-60 & 270 & - \\
    & 60-70 & 307 & - \\
    & 70-80 & 213 & - \\
    & 80-90 & 137 & - \\
    & Not provided & 256 &  200 \\

    Fitzpatrick phototype &
    II or III & - & 200\\ 
    & Not provided & 2000 & 0\\
    \hline
    Expert annotations & Asymmetry & 0 & 200\\
    & Border & 0 & 0 \\
    & Color & 0 & 200 \\
    \hline
    \end{tabular}
\end{table}
\endgroup\label{sec:datasets}

\subsection{Annotations}
We collected annotations from three types of sources: 
\begin{enumerate}
  \item Students of the undergraduate biomedical engineering program at TU Eindhoven,
  \item Crowd workers on the Amazon Mechanical Turk platform, and
  \item Classical image processing algorithms.
\end{enumerate}

Each source assessed the ABC attributes \citep{Abbasi2004-og} as are commonly used by dermatologists: A for asymmetrical shape, B for border irregularity, and C for color of the assessed lesion.

\begin{figure}[ht] 
  \centering
	\includegraphics[width=0.9\textwidth]{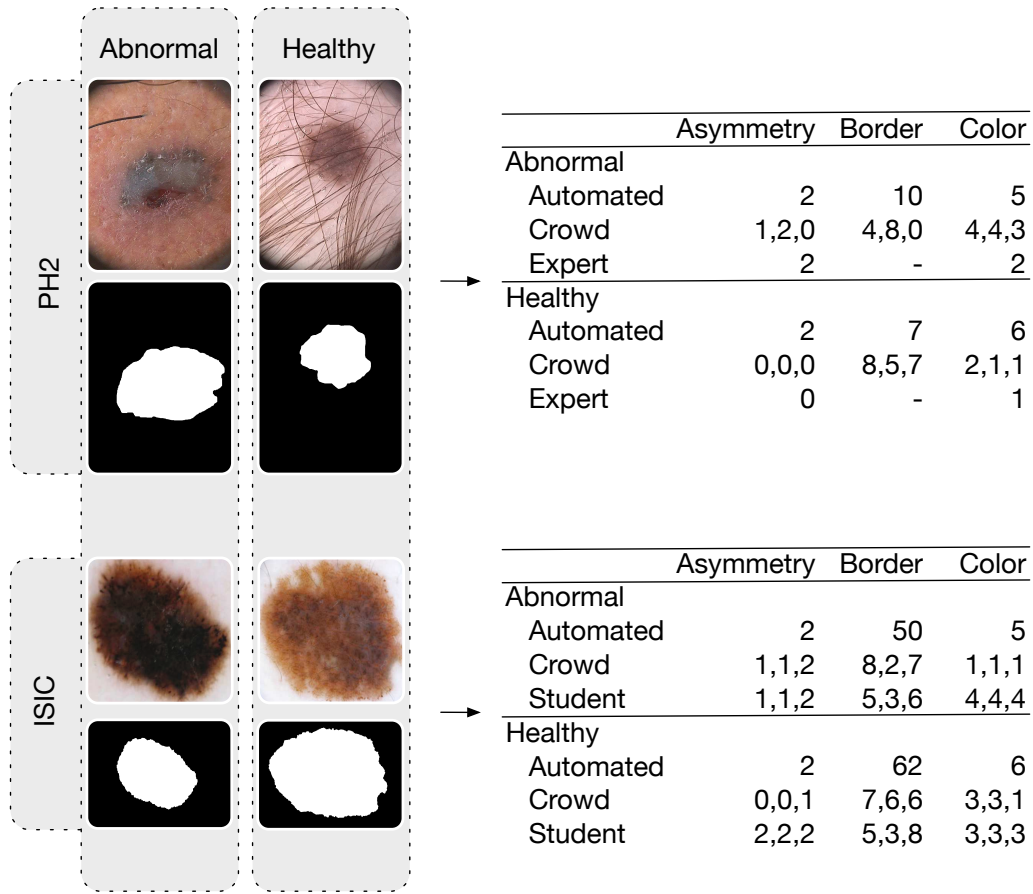}
	\caption{The left part shows examples of skin lesions and corresponding masks (black-white image). The top two lesions are from the PH2 dataset; the bottom two are from the ISIC dataset. The right part shows examples of scores from different annotation sources are also given. The student and crowd annotations indicate that at least three different annotators assessed each image.}
	\label{VISUALISATION_LESIONS}
\end{figure}

When training algorithms, multiple annotations per image were combined as follows: we normalized the annotation scores per annotator (resulting in standardized
data with mean 0 and standard deviation 1) and then averaged these standardized scores per lesion. Early experiments by \citep{Cheplygina2018-ee} showed that this was a justifiable strategy.

\subsubsection{Student annotations}
Undergraduate students without prior knowledge of rating dermoscopic images gathered annotations following a similar protocol as in \citep{Cheplygina2018-ee}, in the context of a medical image analysis course. Each group annotated visual features of 100 images. The students were not blinded to the diagnostic labels, as ISIC is a public dataset. Each group could decide which visual features they wanted to focus on, and the type of annotation scale they wanted to use. However, each visual feature had to be annotated by at least three students from the group. All groups annotated at least some of the ABC features, resulting in 1631 images with asymmetry, border, and color annotations. As we only included the PH2 dataset in the current study after this educational activity was finished, the students did not annotate images from the PH2 dataset.

\subsubsection{Crowd annotations}
Crowd workers on Amazon Mechanical Turk received instructions of how to rate the ABC features based on three examples. Figures \ref{fig:mturk_instructions}, \ref{fig:mturk_examples} and \ref{fig:mturk_annotation} in (\nameref{appendix:amazon_mturk}) illustrate how the annotation task is presented to crowd workers.  The annotation scales were as follows: 

\begin{itemize} 
\item Asymmetry: Zero for symmetrical, one of half-symmetrical, two for asymmetrical.

\item Border: The degree of irregularities in the skin lesion's edge for the border score, with a maximum score of eight. A score of eight indicates irregularities in the entire edge, a score of four in the half-edge, and a zero score indicates no irregularities.

\item Color: The number of the following colors being present in the lesion: light brown, dark brown, white, blue-gray, black, and red.
\end{itemize}

We requested three different crowd workers to annotate each image. The diagnostic labels were not provided, but due to the public nature of the dataset, could be potentially be known. Only workers who had done more than 500 approved tasks previously with an approval rate higher than 90\% could participate. We paid the workers \textdollar0.05 per task. The workers annotated 1250 lesions from ISIC, and all 200 lesions from PH2 with asymmetry, border, and color annotations.

\subsubsection{Automated annotations}

We used classical image processing algorithms from the literature (\cite{Kasmi2016-if}, \cite{Jaworek-Korjakowska2015-fp} and \cite{Achanta2012-zk}) to automatically extract asymmetry, border and color features. We implemented these algorithms based on the papers, as code was not provided. We explain the different steps of the automated asymmetry, border and color algorithms in \nameref{appendix:automatic_annotations}.

Note that we used the provided segmentation masks, and we observed a wide variation of mask boundary definitions: ranging from precisely defined to more loose boundary ones, including clipped boundaries, therefore not every algorithm could be applied to all images. Table \ref{tab:data_summary_annotations} summarizes the number of collected annotations for each annotation source and type.

\begingroup
\setlength{\tabcolsep}{10pt}
\begin{table}[ht]
    \caption{Summary of collected annotations.}
    \vspace{5mm}
    \centering
    \begin{tabular}{lcccccc}
    \hline
    &  \multicolumn{3}{c}{ISIC}  &  \multicolumn{3}{c}{PH2}\\
    \hline
    Automated &&\\
    \hspace{3mm}Asymmetry (\autoa) & 1970\footnotemark[1] & 165\footnotemark[1]\\
    \hspace{3mm}Border (\autob) & 1996\footnotemark[1] & 0\\
    \hspace{3mm}Color (\autoc) & 2000 & 200\\
    \hline
    Crowd &&\\
    \hspace{3mm}Asymmetry (\crowda) & 1250 & 200\\
    \hspace{3mm}Border (\crowdb) & 1250 & 200\\
    \hspace{3mm}Color (\crowdc) & 1250 & 200\\
    \hline
    Student &&\\
    \hspace{3mm}Asymmetry (\studa) & 1631 & 0\\
    \hspace{3mm}Border (\studb) & 1631 & 0\\
    \hspace{3mm}Color (\studc) & 1631 & 0\\
    \hline
    Expert &&\\
    \hspace{3mm}Asymmetry (\expa) & 0 & 200\\
    \hspace{3mm}Border (\expb) & 0 & 0\\
    \hspace{3mm}Color (\expc) & 0 & 200\\
    \hline
    \end{tabular}
    \label{tab:data_summary_annotations}
\end{table}
\endgroup\label{sec:annotations}

\section{Experiments}
To investigate the added value of our annotations, we address a binary classification problem: healthy (nevi) vs abnormal (melanoma and seborrheic keratosis). 

\subsection{Annotation analysis}

To gain insight into the characteristics of the collected annotations, we use visual representations that provide detail and overview at the same time. In particular, we created raincloud plots \citep{Allen2019-vd} with A, B, and C annotations for each annotation source.  

We compared the agreement level between the three types of annotation sources. We first averaged the image's scores separately for A, B, and C annotations and then standardized these mean scores, resulting in z-scores. Based on the z-scores, we calculated the agreement level using Pearson correlations. 

To quantitatively measure how related the annotation sources and the diagnostic labels (abnormal and healthy) are, we computed the Pearson's correlation coefficient between them for all collected annotations for the ISIC and the PH2 dataset.

We interpret the Pearson's correlation coefficient $\rho$ as follows: weak correlation for $0 \leq \rho < 0.3$, moderate correlation for $0.3\leq \rho < 0.5$, strong correlation for $0.5 \leq \rho < 1$.

\subsection{Multi-task learning}
\begin{figure}[ht]
    \centering
    \includegraphics[width=0.9\textwidth]{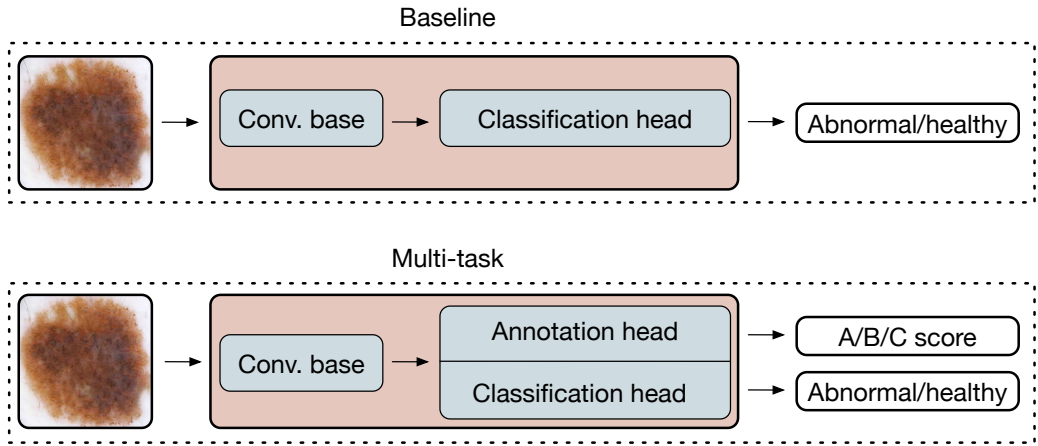}
    \caption{Architecture of baseline (top) and MTL (bottom) models. All models are on top of the VGG-16, Inception v3 or ResNet50 convolutional base.}
    \label{base_and_mtl}
\end{figure}

To understand how the collected annotations can contribute to training deep learning algorithms, we performed several experiments where we compared a baseline model which is trained only on the diagnosis, and a MTL model which is trained on both the diagnosis and one or more of the collected ISIC based ABC annotations (Fig.~\ref{base_and_mtl}). We compared these for three different CNN architectures: VGG-16 \citep{simonyan2015deep}, Inception v3 \citep{Szegedy2016-me}, and ResNet50 \citep{He2016-ow}.

The \textbf{baseline model} extends the convolutional base with two fully connected layers with a sigmoid activation function. During training, we use class weights to pay more attention to samples from the underrepresented class. We use cross-entropy as a loss function. All choices for the baseline model were made based on only the training and validation set. We fine-tune the baseline model until the performance on the validation dataset was within the performance of the ISIC 2017 challenge. 

The \textbf{multi-task model} extends the convolutional base with three fully connected layers. The model has two outputs with different network heads: one head is the classification output, the other represents the visual feature (such as asymmetry). We use a customized mean squared error loss \( L_{MTL} \) and a last layer linear mapping for the annotation-regression task. We used a vector \( a_i \) in which we store per lesion whether or not an annotation is available. If an annotation is present, then \( a_i \) is equal to 1; otherwise, \( a_i \) is 0. This vector is used in a custom loss function that calculates the mean squared error. In case no annotation is available, it does not affect the weights, and the error is not propagated. For the binary classification task, we used a cross-entropy loss \( L_c \) and sigmoid activation function of the nodes. The contribution of the losses is equal -- we did not optimize the weighting parameter, but early experiments showed that the contribution of the classification loss should not be too low. The resulting loss values are summed and minimized during network training. For our MTL models, the loss function (\ref{eq:loss}) is as follows:

\begin{align}
L_{MTL}=0.5({\frac 1 A} {\sum_{i=1}^N} a_i(y_i-\hat{y_i})^2) + 0.5L_c
\label{eq:loss}
\end{align}

where \(A\) denotes the number of available annotations, \(N\) represents the number of lesions, and \(y_i\) and \(\hat{y_i}\) denotes respectively the prediction and the expected outcome for lesion \(i\).

To investigate the value of the additional annotations, we compare baseline networks with multi-task networks to which we add one of the following nine types of annotations individually: 
\begin{itemize}
    \item Asymmetry: \autoa, \crowda, \studa
    \item Border: \autob, \crowdb, \studb
    \item Color: \autoc, \crowdc, \studc
\end{itemize}

We also combine the networks above into ensembles, by averaging the network predictions with equal weights (earlier experiments \citep{Raumanns2020-ac} showed limited effectiveness of optimizing the weights). We tested the following three ensembles:
\begin{itemize}
    \item StudA + StudB + StudC (named studABC)
    \item CrowdA + CrowdB + CrowdC (named crowdABC)
    \item AutoA + AutoB + AutoC (named autoABC)
\end{itemize}

Finally, we trained the ResNet50 MTL models with synthetic annotations. We replaced the original A, B and C scores with uniformly randomized ones in the range [0.0, 1.0] and combined them into ensembles.

The experimental procedure is as follows: we trained all CNN models using 70\% as training, 17.5\% as validation and 12.5\% as test proportions, in a five-fold cross-validation approach. 
We trained all layers, including the convolutional base, starting with the pre-trained ImageNet weights. For training, we used 30 epochs with a batch size of 20 using the default backpropagation algorithm RMSprop \citep{Tieleman2012-kp} as the optimizer, with a learning rate of $2.0\mathrm{e}{-5}$. 

To determine a reasonable learning rate, we evaluated six learning rates on the baseline VGG-16 model (varying over 2 orders of magnitude, between 1.0e-3 and 1.0e-5). We did not vary the learning rate across the different model architectures to keep the number of experiments reasonable while still having a systematic comparison. A description of all hyperparameters is available in Table \ref{tab:hp_table} (\nameref{appendix:hyperparamters}).

We performed the ensemble of the predictions of multi-task models through averaging; we based the lesion's classification on the predictions of the three multi-task models, with each model's prediction having an equal weight (a third). 

We compared the average area under the receiver operating characteristic curve (AUC) of the baseline model to the average AUC scores of the different multi-task models and ensembles. The average AUC score was calculated per experiment, taking the average of the AUC score of each fold.

We realized our deep learning models in Keras using the TensorFlow backend \citep{Geron2019-rs}.

\section{Results}

\subsection{Correlations between annotations and the diagnostic label}

\begin{figure}[ht]
    \centering
    \includegraphics[width=0.8\textwidth]{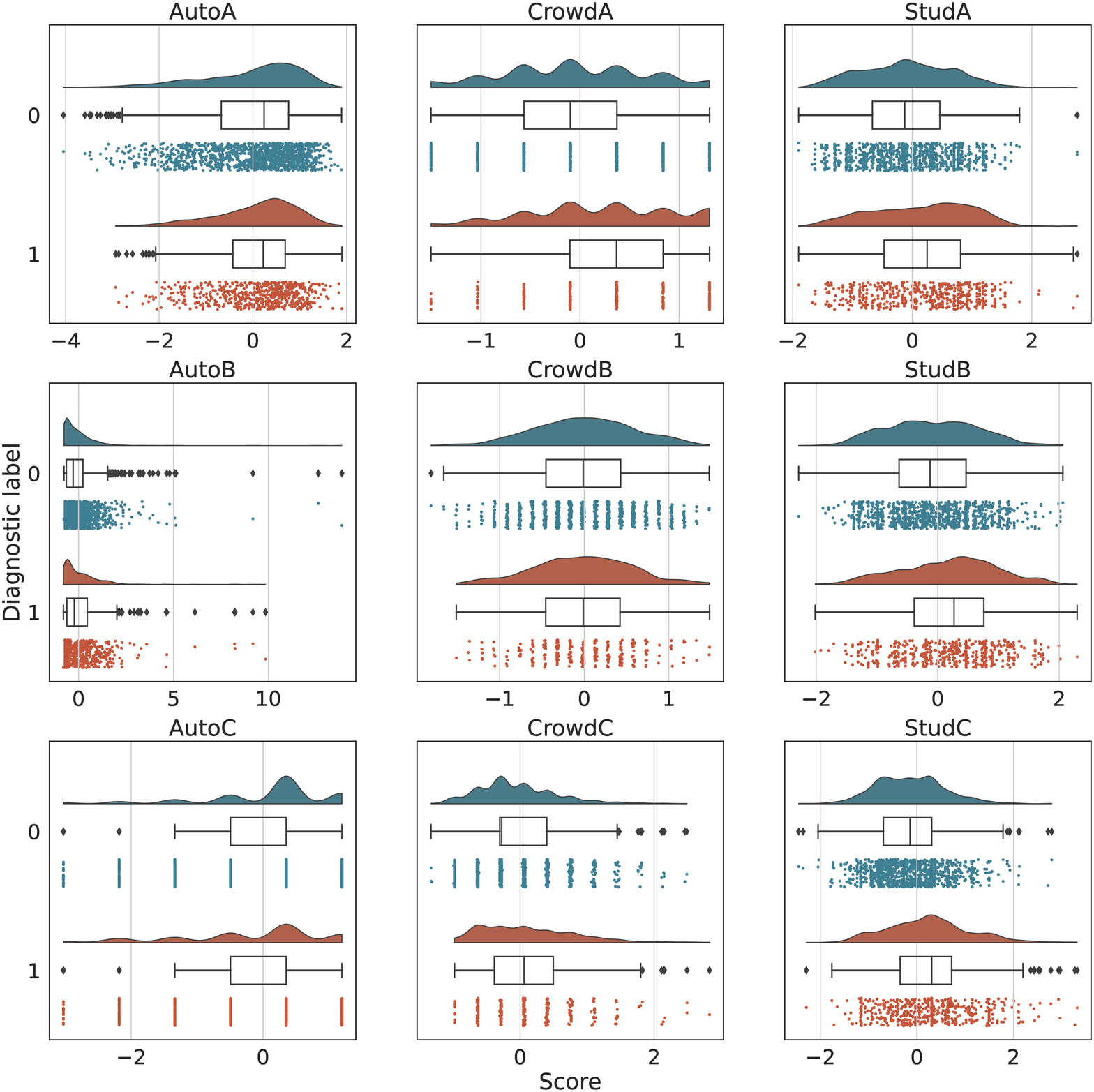}
    \caption{Raincloud plots of automated, crowd, and student ISIC annotations. Diagnostic label 1 stands for abnormal, label 0 for healthy lesions.}
    \label{fig:rainclouds_isic}
\end{figure}

\begin{figure}[ht]
    \centering
    \includegraphics[width=0.8\textwidth]{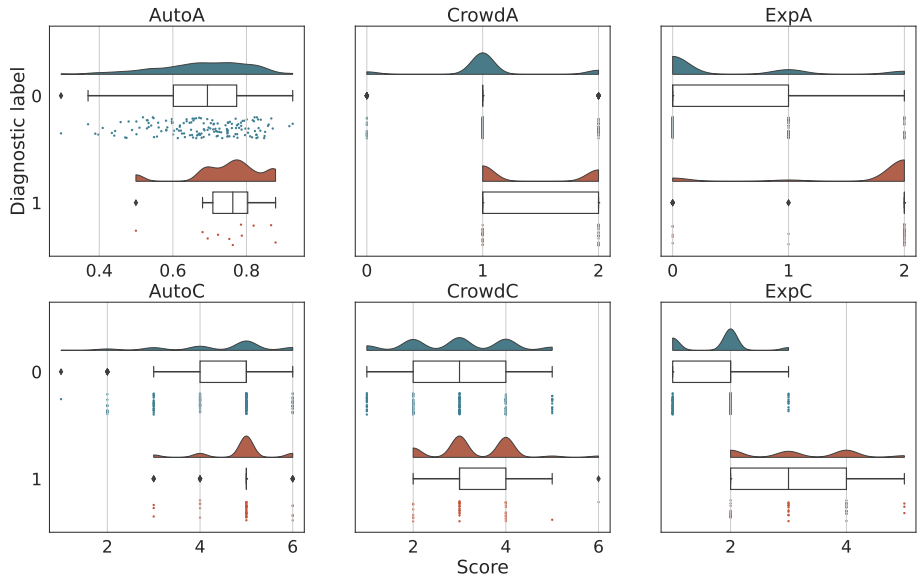}
    \caption{Raincloud plots of automated, crowd, and expert PH2 annotations. Diagnostic label 1 corresponds to abnormal lesions.}
    \label{fig:rainclouds_ph2}
\end{figure}

The first research question is whether annotations for ABC criteria are discriminative for diagnosis. We show the distributions of asymmetry, border and color versus the label (normal or abnormal) for each annotation source in Fig. \ref{fig:rainclouds_isic} (ISIC) and Fig. \ref{fig:rainclouds_ph2} (PH2). Table \ref{tab:pearson_table_annotations_diagnostic_label} lists Pearson correlation coefficient measures between annotation type and diagnostic label for ISIC annotations.


\begin{table}[ht]
    \centering
    \caption{Pearson correlation coefficient measures between annotation type and diagnostic label}
    \begingroup
\setlength{\tabcolsep}{10pt}
\begin{tabular}{lrrr}
\hline
      Type &      A &       B &       C \\
\hline
 Automated & 0.046 &  0.071 & -0.122 \\
     Crowd & 0.166 & -0.017 &  0.103 \\
  Student & 0.151 &  0.161 &  0.230 \\
\hline
\end{tabular}
\endgroup

    \label{tab:pearson_table_annotations_diagnostic_label}
\end{table}

 The distributions of the A, B and C scores for each diagnostic label overlap considerably. \studa, \studb, \studc, \crowda, and \crowdc{} sample medians for abnormal lesions were greater than sample medians for normal ones. \autoa, \autob, \autoc, and \crowdb{} sample medians for abnormal and normal lesion are almost identical, although the underlying nature of raw \autoa{} observations is different. The width of the \autoa{} annotations belonging to the normal distribution is greater than those of the abnormal distribution.

The data of ExpA are right-skewed for normal lesions and left-skewed for abnormal ones; their normal and abnormal distributions do not overlap. CrowdA for normal lesions is unimodal; the data is centered around the middle. CrowdA for abnormal lesions is bimodal. The CrowdA's abnormal and normal distributions partially overlap. AutoA's distributions for abnormal and normal lesions overlap considerably, and the sample median for abnormal is greater than the sample median for normal ones. 

The distributions of ExpC, CrowdC and AutoC are multimodal. The ExpC's abnormal and normal distributions partially overlap. CrowdC's and AutoC's distribution for abnormal and normal lesions considerably overlap.

The Pearson's coefficient (Table \ref{tab:pearson_table_annotations_diagnostic_label}) indicates there is no correlation between the crowd-B annotations and the diagnostic label. No correlation also applies to the \autoa{} and \autob{} annotations. We see a weak correlation between all three types of student annotations (\studa, \studb, and \studc), the \autoc{} annotations and the diagnostic label.

\subsection{Inter agreement level between the different annotation sources}

\begin{figure}[h!]
    \begin{tabular}{@{}c@{}}
    \includegraphics[width=0.3\textwidth,height=4cm]{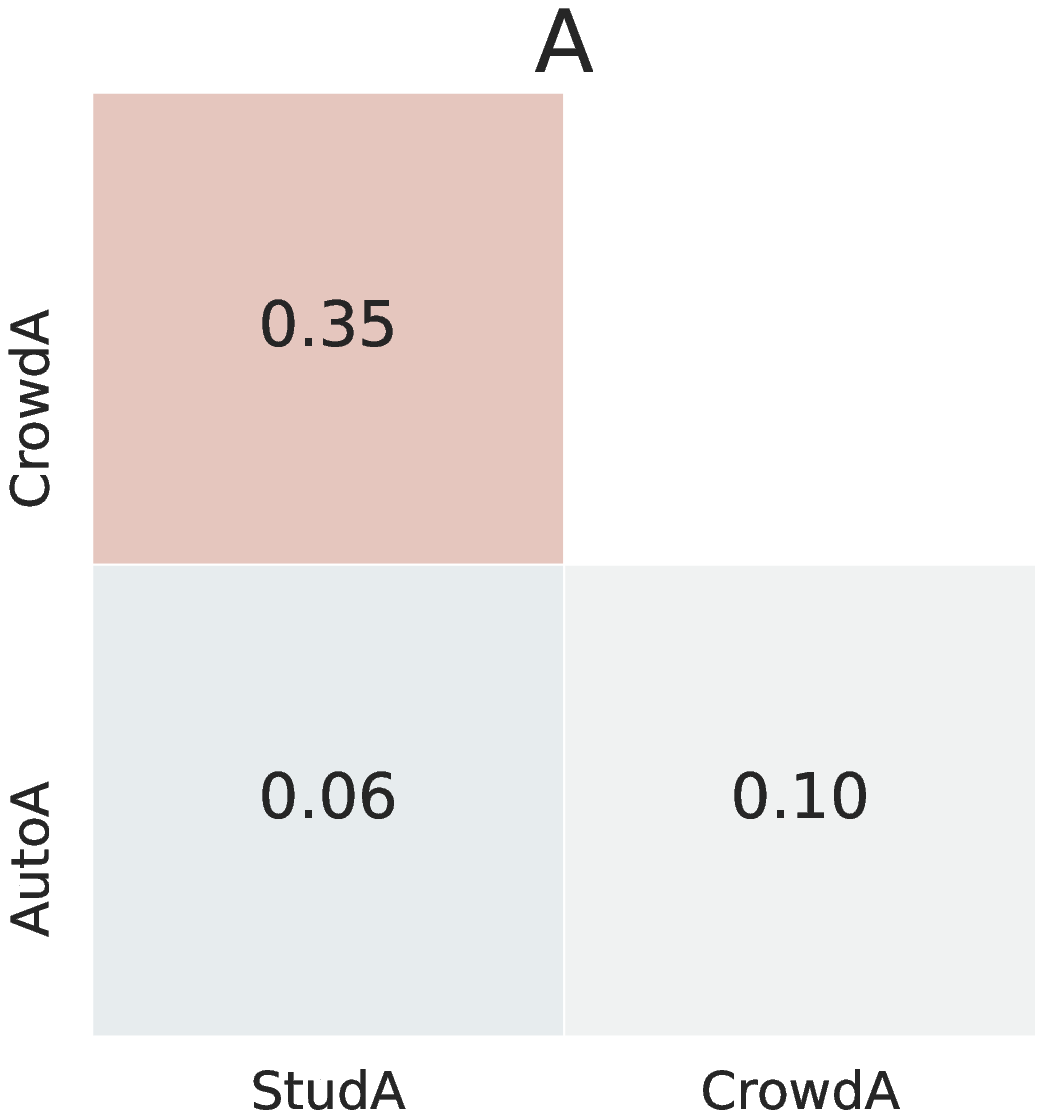}
  \end{tabular}
  \begin{tabular}{@{}c@{}}
    \includegraphics[width=0.3\textwidth,height=4cm]{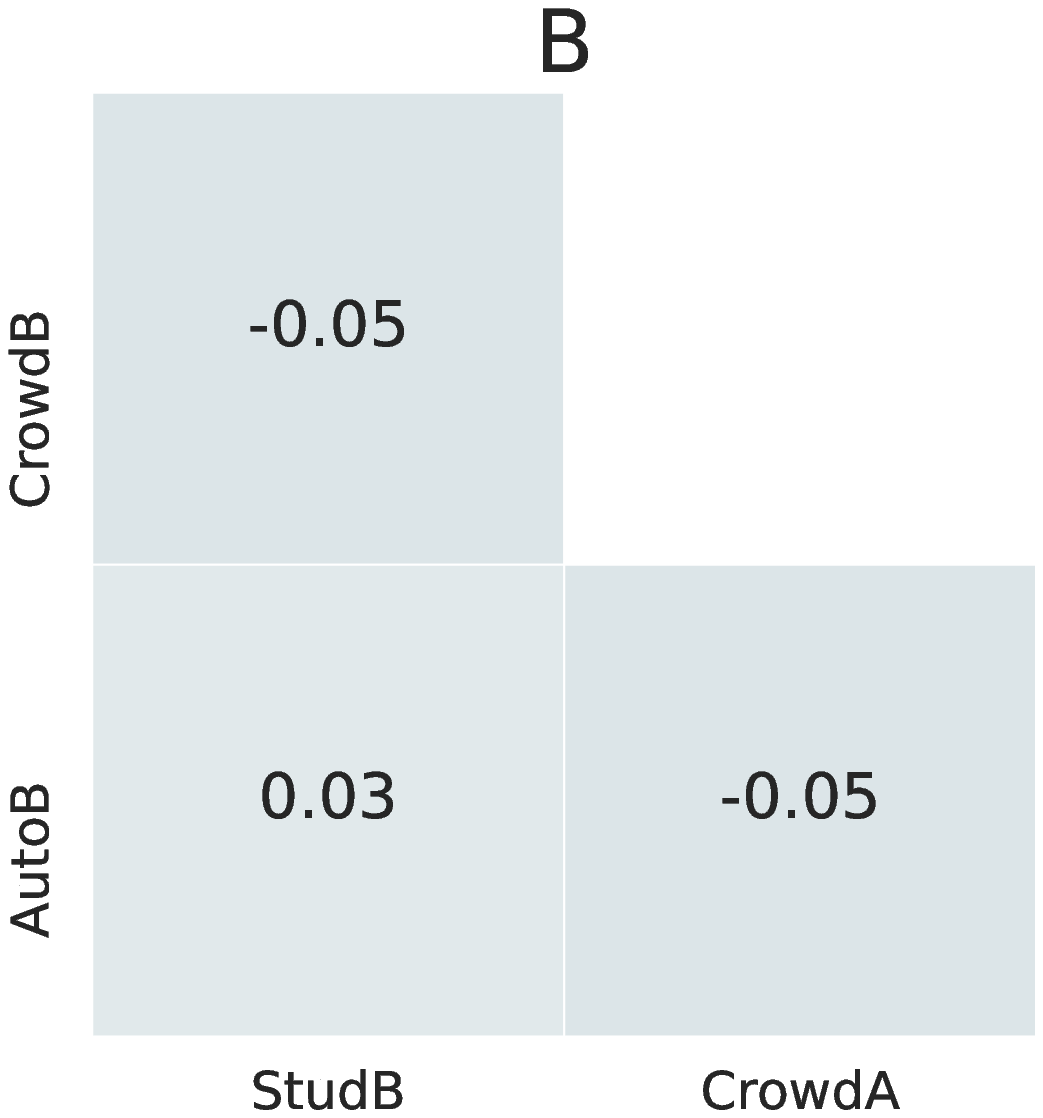}
  \end{tabular}
  \begin{tabular}{@{}c@{}}
    \includegraphics[width=0.3\textwidth,height=4cm]{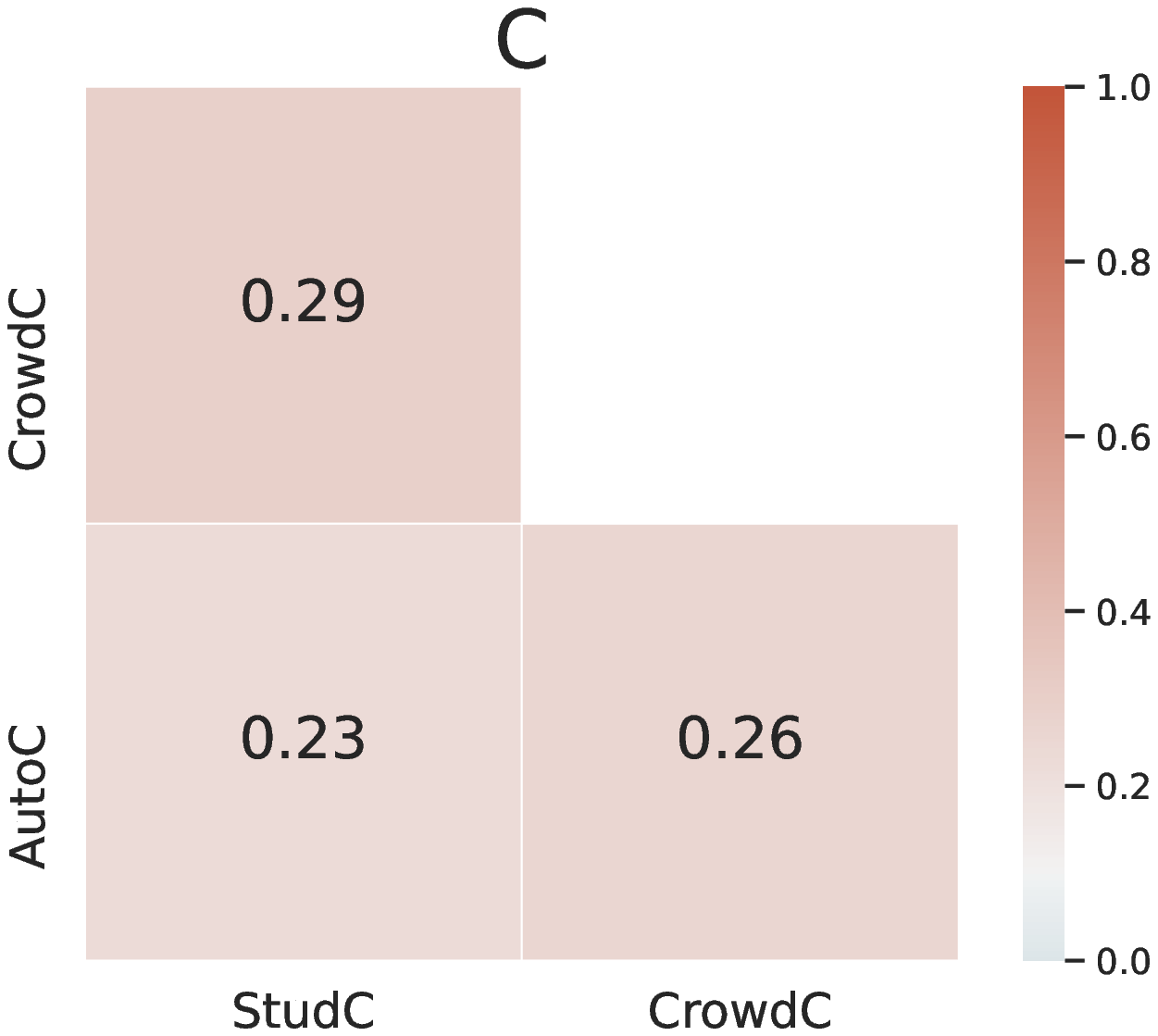}
  \end{tabular}
  \caption{Pearson's correlation coefficient measures for A, B and C features}
  \label{fig:pearson}
\end{figure}

The second research question examined the level of agreement between annotation sources for the A, B and C criteria. In order to answer this question, the results of the analysis of agreement level using Pearson correlations were examined. Figure \ref{fig:pearson} illustrates these correlations graphically.

For asymmetry, we observed a moderate positive correlation between student and crowd annotations. For color, we observed a weak correlation between all annotation sources. We observed no correlation between the different sources for border annotations.

\subsection{Effect on CNN performance for multiple annotations usage}

Our last research question examines the performance benefit of using multiple annotations in a multi-task model. We used the annotations together with skin lesion images for training multi-task CNNs. We explored how they affect the model’s performance compared to a baseline model without using annotations.

\begin{figure}[h!]
    \centering
    \includegraphics[width=0.9\textwidth]{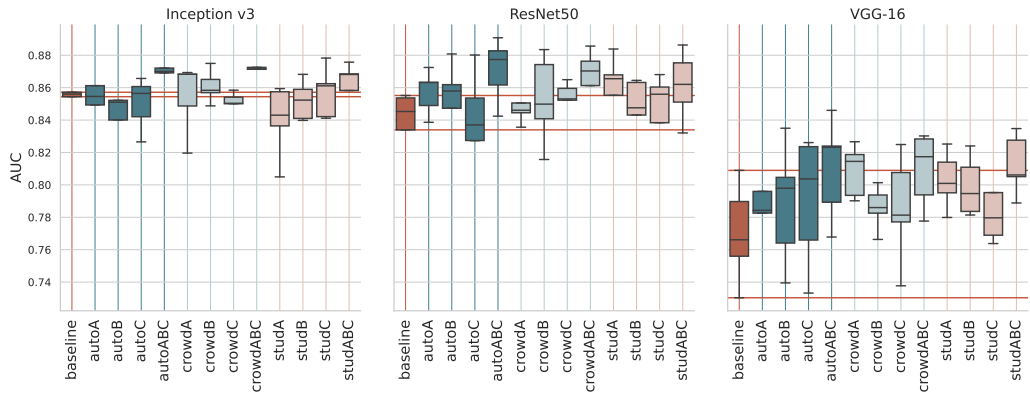}
    \caption{AUCs of baseline and multi-task models (asymmetry, border and color) per CNN architecture.}
    \label{fig:auc_combined}
\end{figure}

The AUCs of the baseline and multi-task models are shown in Fig. \ref{fig:auc_combined} (the numerical results are also available in \nameref{appendix:auc_performances}). Table \ref{tab:ensembles_auc} compares the AUCs of the ensembles based on randomized annotations and original annotations.

\begin{table}[ht]
    \centering
    \caption{Mean AUCx100 performances of ResNet50 MTL ensembles trained with original and randomized annotations}
    \setlength{\tabcolsep}{10pt}
\begin{tabular}{llll}
\hline
Annotations &      autoABC &     crowdABC &      studABC \\
\hline
   Original &  87.1 ± 1.9  &  86.3 ± 2.4  &  86.1 ± 2.1  \\
 Randomized &  85.0 ± 2.4  &  84.4 ± 2.8  &  84.6 ± 2.4  \\
\hline
\end{tabular}
    \label{tab:ensembles_auc}
\end{table}

The AUC values for each CNN architecture are close together. The Inception v3 CNN outperformed the ResNet50 and VGG-16 architectures for all three features, but for studA, autoA and autoB multi-task models, ResNet50 performs slightly better. The three ensembles show improved performance when compared to the other models in the same architecture. The best performing model is the autoABC ResNet50 ensemble model. For VGG-16 and ResNet50 the multi-task models outperform the baseline models. For Inception v3 only studC, studABC, crowdA, crowdB, crowdABC, autoA and autoABC models outperform the baseline model. Table \ref{tab:ensembles_auc} shows that ResNet50 MTL ensembles trained with retrieved annotations perform better than ensembles with randomized annotations.

\section{Discussion}
\subsection{Summary of results}

\subsubsection{At most moderate correlations between annotations and the diagnostic label}
In general, the acquired non-expert annotations (especially the student annotations) show a weak correlation with the diagnostic label (Table \ref{tab:pearson_table_annotations_diagnostic_label}). At the same time, automated annotations show only a correlation for the color feature. Results in a study by \cite{Giotis2013-iz} support this. They show that the color feature has a discriminative power, but the border feature not. In their study, they did not assess asymmetry.

\subsubsection{Weak to moderate inter agreement level between the different annotation sources}
We observed low agreement between student and crowd for asymmetry and color scores and comparable agreement level between automated and non-expert color features. In contrast, we observed disagreement for the border feature between non-experts and automated scores. Based on these agreement levels, annotating the asymmetry and color seems more straightforward than annotating the border.

Measures with disagreement can still be informative. According to Cheplygina et al., "Although removing disagreement leads to the best performances, we show that disagreement alone leads to better-than-random results. Therefore, the disagreement of these crowd labels could be an advantage when training a skin lesion classifier with these crowd labels as additional outputs" \cite{Cheplygina2018-ee}. 

\subsubsection{A mainly positive effect on CNN performance for multiple annotations usage}
Our preprint \citep{Raumanns2020-ac} showed that VGG-16 based multi-task models with individual crowdsourced features have a limited effect on the model. Still, when combined in ensembles, it leads to improved generalization. This paper has further confirmed this improved generalization for Inception v3 CNN and the ResNet50 based multi-task models. We also showed that the Inception v3 CNN outperformed the ResNet50 and VGG-16 architectures for all three annotations. The ResNet50 showed improved performance compared to the VGG-16 model. 

One perhaps surprising result is that noisy annotations positively contribute to the performance. Our annotations are highly noisy, as shown by the correlation coefficients in Table \ref{tab:pearson_table_annotations_diagnostic_label} and the raincloud plots in Fig. \ref{fig:rainclouds_isic}. We explain the success of these annotations by several factors. First, annotations that are perfectly correlated with the diagnostic label would not add information to the classifier. Second, training multiple (biased) classifiers and combining these in an ensemble is a well-known result that can lead to improved generalization \citep{jain2000statistical}. We also show that this is the case for fully randomized annotations (Table \ref{tab:ensembles_auc}), but that this is less effective than our gathered annotations. Finally, the classifier also learns from the diagnostic labels; learning from only the annotations is not feasible.

\subsection{Limitations}
We demonstrated the added value of using non-expert and automated annotations for model learning. Yet, we have not optimally tuned the algorithms for automatically scoring the ABC features. Also, the algorithms depend on the available masks. The quality of these masks differs and affects the automated score of asymmetry and border irregularity. Furthermore, we did not optimize the multi-task model to achieve the highest performance.

There are three main differences between the annotation by the crowd and the students. Firstly, the student annotations were done in an educational setting, and students were encouraged to decide their annotation protocols based on existing literature.  Letting the crowd decide on their protocol was not feasible due to the distributed nature of crowdsourcing: the crowd workers do not interact with each other, and most annotate only a few images.  Secondly, students were not blinded to the diagnostic label while annotating. Since it is a public dataset, anybody could try to find the label, but this is less likely for MTurk because the MTurk jobs are time-limited. Thirdly, only the ISIC skin lesions were part of the student assignment. The PH2 dataset was only added to this study after an informal review round after the subject was no longer taught.

\cite{Thompson2016-ji} conclude that personal characteristics (such as sex and age) influence healthcare-seeking behavior; thus, the datasets might have an unintended bias against subjects from specific groups. \cite{Abbasi-Sureshjani2020-hn} showed that even when the training set is relatively balanced, there is a variable performance on subgroups based on age and sex. For the ISIC dataset, the distribution between men and women is roughly the same. However, the dataset is skewed towards lower age groups. Unfortunately, the sex and age characteristics are missing for the PH2 dataset. Besides these, the Fitzpatrick phototype \citep{Sachdeva2009-pj} is another characteristic that would be relevant in this prediction problem. The PH2 dataset contains lesions with phototype II or III \citep{Mendonca2015-uc}, but it is unspecified per lesion, and the ISIC database does not have phototype information. The discussed unintended bias or missing demographic characteristics might lead to representation bias, ultimately leading to an insufficiently representative model.

\subsection{Opportunities}

In this study, we used automated annotations based on best in class image algorithms. However, the automated annotations are not as discriminative as the expert annotations for diagnosis. A future study could concentrate on tuning the used image algorithms or finding new algorithms that mimic expert annotations more closely.

Another important direction for future research is explainability. We showed that multi-task models with non-expert annotations have a limited effect on the model performance. However, the multi-task prediction of ABC scores together with a diagnosis helps to explain the diagnostic outcome. For example, when the model predicts abnormality and high ABC scores, the latter could be interpreted as warning signs/explanation for abnormality. The use of the model in this way needs further research. For instance, what kind of model architectures fit the best for this explanation task? What types of annotations are suitable for this task? What kind of model produces gradual class activation maps \citep{Selvaraju2017-ff} that explain the model outcome to dermatologists?

We used three different sources to collect annotations for the ISIC and PH2 dataset to create a new open dataset that incorporates these additional annotations and then used this dataset to see whether a multi-task model can take advantage of the extra annotations. A future study, for example, might consider clarifying the multi-task model performance benefit of additional annotations for mammography in the domain of breast cancer diagnosis or studying the quality of the extra annotations in another field.

The annotators' expertise levels will probably be different. Depending on the level of expertise, an annotation can have more or less influence during the model's training procedure. This could be possible via different weights or by modeling the expertise of each annotator, similar to \cite{guan2017who}. However, this might be more applicable for few annotators who annotated all images, rather than many annotators who each annotate a few images, as in crowdsourcing.

This study used the ground truth masks’ segmentation masks for extracting asymmetry, border irregularity, and color characteristics. We see a wide variation in the given segmentation masks, with boundaries ranging from precisely defined to more loose ones. Some masks also have clipped boundaries. Optimizing these masks need to be considered in future studies.

\section{Conclusions}
This paper presented ENHANCE, an open dataset with additional annotations, gathered by three different annotation sources, for ISIC 2017 and PH2 images of skin lesions, describing the lesions' asymmetry, border, and color. We tested the dataset extensively with three different network architectures (VGG-16, ResNet50 and Inception c3) and show that the often claimed enhancement of multiple tasks is marginal.

The asymmetry and color annotations from different annotation sources show their discriminative power to diagnose healthy or abnormal. At the same time, there is a moderate agreement level between the annotation sources. Nevertheless, this disagreement could be advantageous when training a skin lesion classifier with the annotations as additional output.


\acks{We would like to acknowledge all students and crowd workers who have contributed to this project with image annotation. We gratefully acknowledge financial support from the Netherlands Organization for Scientific Research (NWO), grant no. 023.014.010.}

%
\ethics{The work follows appropriate ethical standards in conducting research and writing the manuscript, following all applicable laws and regulations regarding treatment of animals or human subjects.}

\coi{We declare we don’t have conflicts of interest.}

\bibliography{refs_ralf, refs_veronika}

\clearpage
\appendix 
\section*{Appendix A. AUC performances}\label{appendix:auc_performances}
Table \ref{tab:aucs_table} shows the mean AUCx100 of 5-fold cross-validation of Inception v3, ResNet50 and VGG-16 CNN baseline and multi-task models. We trained the multi-task models on all features (A, B and C) from three different annotation sources (student, crowd, and automated). The AUC of the best performing model per CNN architecture is emphasized in bold.

\begin{table}[h]
    \centering
    \caption{Mean AUCx100 performances of 5-fold cross-validation on all features (\textbf{A}symmetry, \textbf{B}order and \textbf{C}olor) for three different CNNs.}
    \setlength{\tabcolsep}{10pt}
\begin{tabular}{llll}
\hline
     Type & Inception v3 &     ResNet50 &       VGG-16 \\
\hline
 baseline &  85.3 ± 1.2  &  83.6 ± 2.7  &  77.0 ± 3.0  \\
    autoA &  85.4 ± 2.3  &  85.7 ± 1.3  &  79.0 ± 2.6  \\
    autoB &  84.8 ± 2.0  &  85.3 ± 2.4  &  78.8 ± 3.7  \\
    autoC &  85.0 ± 1.6  &  84.5 ± 2.2  &  79.1 ± 4.0  \\
  autoABC &  86.4 ± 1.8  &  \textbf{87.1} ± 1.9  &  81.0 ± 3.1  \\
   crowdA &  85.5 ± 2.2  &  84.8 ± 1.0  &  80.9 ± 1.6  \\
   crowdB &  86.1 ± 1.0  &  85.3 ± 2.7  &  78.6 ± 1.3  \\
   crowdC &  85.1 ± 0.6  &  84.2 ± 3.5  &  78.6 ± 3.3  \\
 crowdABC &  \textbf{86.7} ± 1.4  &  86.3 ± 2.4  &  80.9 ± 2.3  \\
     studA &  84.0 ± 2.2  &  85.8 ± 2.6  &  80.3 ± 1.7  \\
    studB &  85.2 ± 1.2  &  84.6 ± 2.1  &  79.9 ± 1.8  \\
    studC &  85.7 ± 1.6  &  84.4 ± 2.7  &  78.9 ± 2.9  \\
  studABC &  86.2 ± 1.4  &  86.1 ± 2.1  &  \textbf{81.2} ± 1.9  \\
\hline
\end{tabular}

    \label{tab:aucs_table}
\end{table}

\clearpage
\section*{Appendix B. Hyperparameters}\label{appendix:hyperparamters}
The table below describes the hyperparameter choices for baseline and MTL experiments.
\begin{table}[ht]
    \centering
    \caption{Describes each parameter, its value, and, when applicable, the name in the Python code}
    \begin{tabular}{llll}
\hline
    Hyperparameter & Value baseline & Value MTL &  Name in code\\
\hline
    Learning rate & $2.0\mathrm{e}{-5}$ & $2.0\mathrm{e}{-5}$ &\\
    Optimizer & RMSprop & RMSprop & \\
    Batch size & 20 & 20 & BATCH$\_$SIZE \\
    Activation function & Sigmoid & Sigmoid (classification) &\\
    & & Linear (annotation) &\\
    Number of epochs & 30 & 30 & EPOCHS\\
    Steps per epoch& 100 & 100 & STEPS$\_$PER$\_$EPOCH\\
    Steps to validate& 50 & 50 & VALIDATION$\_$STEPS\\
    Steps to yield from pred.generator& 20 & 20 & PREDICTION$\_$STEPS\\
    Class weights &balanced &balanced &\\
    Input shape &(384, 384, 3) &(384, 384, 3) &INPUT$\_$SHAPE\\
    Include top &False &False &\\
    Pre-training &Imagenet &Imagenet &\\
\hline
\end{tabular}
    \label{tab:hp_table}
\end{table}

\section*{Appendix C. Automatic annotation}\label{appendix:automatic_annotations}
\begin{figure}[ht]
    \centering
    \includegraphics[width=0.9\textwidth]{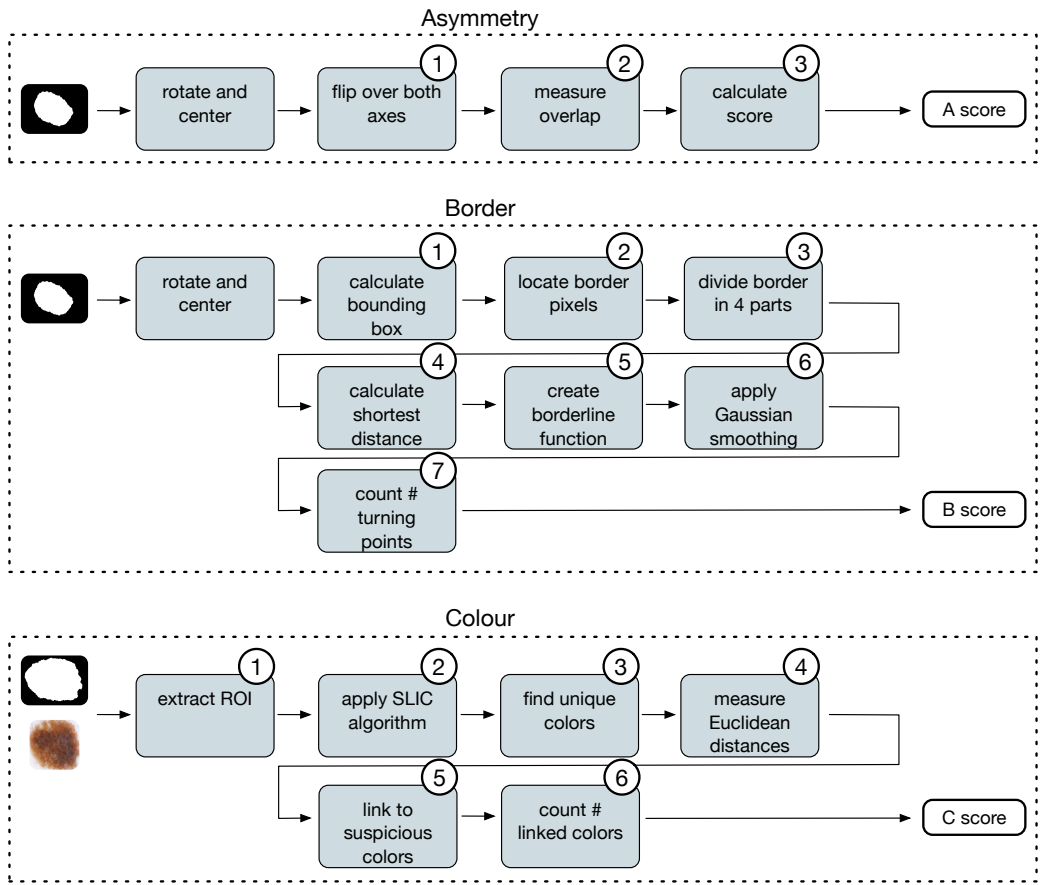}
    \caption{A schematic overview of the different steps of the automated annotation of asymmetry, border and color.}
    \label{fig:automation_steps}
\end{figure}

This appendix explains the different steps of automated asymmetry, border and color algorithms. Fig. \ref{fig:automation_steps} presents a schematic overview of the algorithm annotation steps.

As a preprocessing step, we first rotate and center the provided segmentation mask, such that the major axis (longest diameter of the lesion) passes horizontally through the segmentation center of mass. This is the major axis, the minor axis is then defined as the diameter perpendicular to the major axis and also passing through the center of mass. 

The asymmetry algorithm is based on shape asymmetry \citep{Kasmi2016-if}. The asymmetry score algorithm steps are:
\begin{enumerate}
    \item flip the rotated image over both axes separately to measure the overlap in pixels between the mask areas on either side of the axis;
    \item determine the shape symmetry ratio for an axis by dividing the overlap in pixels by the total amount of pixels in the lesion's mask;
    \item calculate the score by averaging the symmetry ratios of both axes.
\end{enumerate}

The border score is based on \citep{Jaworek-Korjakowska2015-fp} and assesses the border irregularity as follows:
\begin{enumerate}
    \item calculate a bounding box around the segmentation and connect four lines between the bounding box's vertices and the center of mass;
    \item locate the border pixels that are on each of the four lines (resulting in a total of four border pixels);
    \item divide the border into four parts with the help of the four pixels (clockwise: top, right, bottom and left part border pixels);
    \item calculate the shortest distance between the border pixel and the edge's image for each of the four parts (the direction in which the distances are calculated varies per part; for the top border pixels in the upward direction, the right part border pixels in the rightward direction, the bottom border pixels in the downward direction and the left border pixels in the leftward direction);
    \item create a borderline function based on the pixel location on the border and the calculated distances;
    \item smooth the borderline function with a Gaussian filter;
    \item count the number of turning points in the smoothed signal, resulting in the actual border score.
\end{enumerate}

The color algorithm is based on \citep{Kasmi2016-if} and computes the number of the suspicious colors present in the lesion: light brown, dark brown, white, blue-gray, black, and red, as follows:  
\begin{enumerate}
    \item extract the region of interest from the lesion image using the mask;
    \item segment the region of interest into superpixels by using the SLIC superpixels algorithm \citep{Achanta2012-zk};
    \item find all unique colors (having different RGB values) that are present in the SLIC superpixels;
    \item measure the normalized Euclidean distance between each unique color in the set of SLIC superpixels and the six predefined values for the suspicious colors;
    \item link a unique color to a suspicious color when its distance is less than a fixed threshold (0.4);
    \item count a linked unique color as a suspicious color when the number of pixels belonging to that unique color exceeds five per cent of the total amount of pixels of the region of interest;
\end{enumerate}

\clearpage
\section*{Appendix D. Amazon MTurk annotation}\label{appendix:amazon_mturk}
\begin{figure}[ht]
    \centering
    \includegraphics[width=0.6\textwidth]{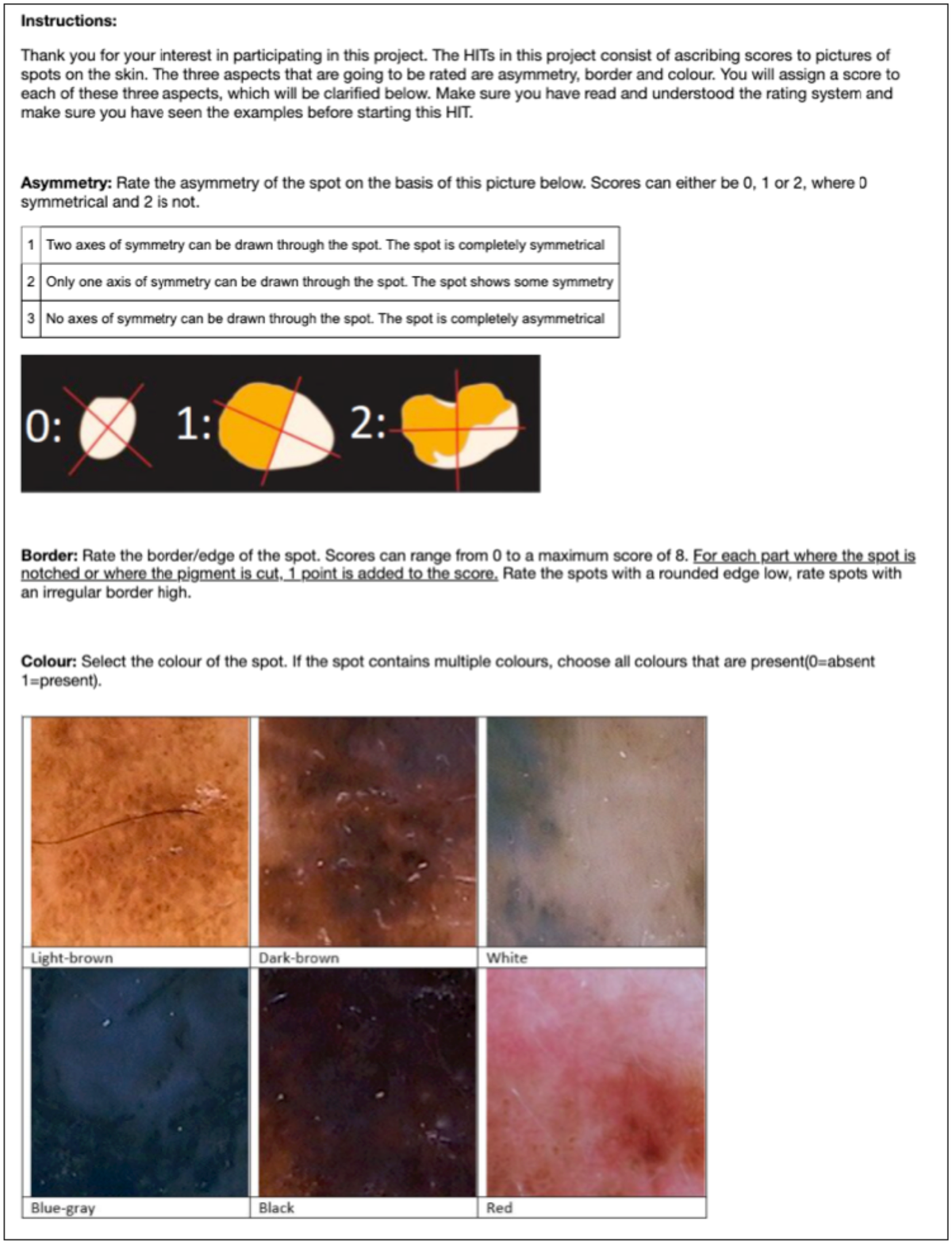}
    \caption{An instruction screen with the task description.}
    \label{fig:mturk_instructions}
\end{figure}

\begin{figure}[ht]
    \centering
    \includegraphics[width=0.6\textwidth]{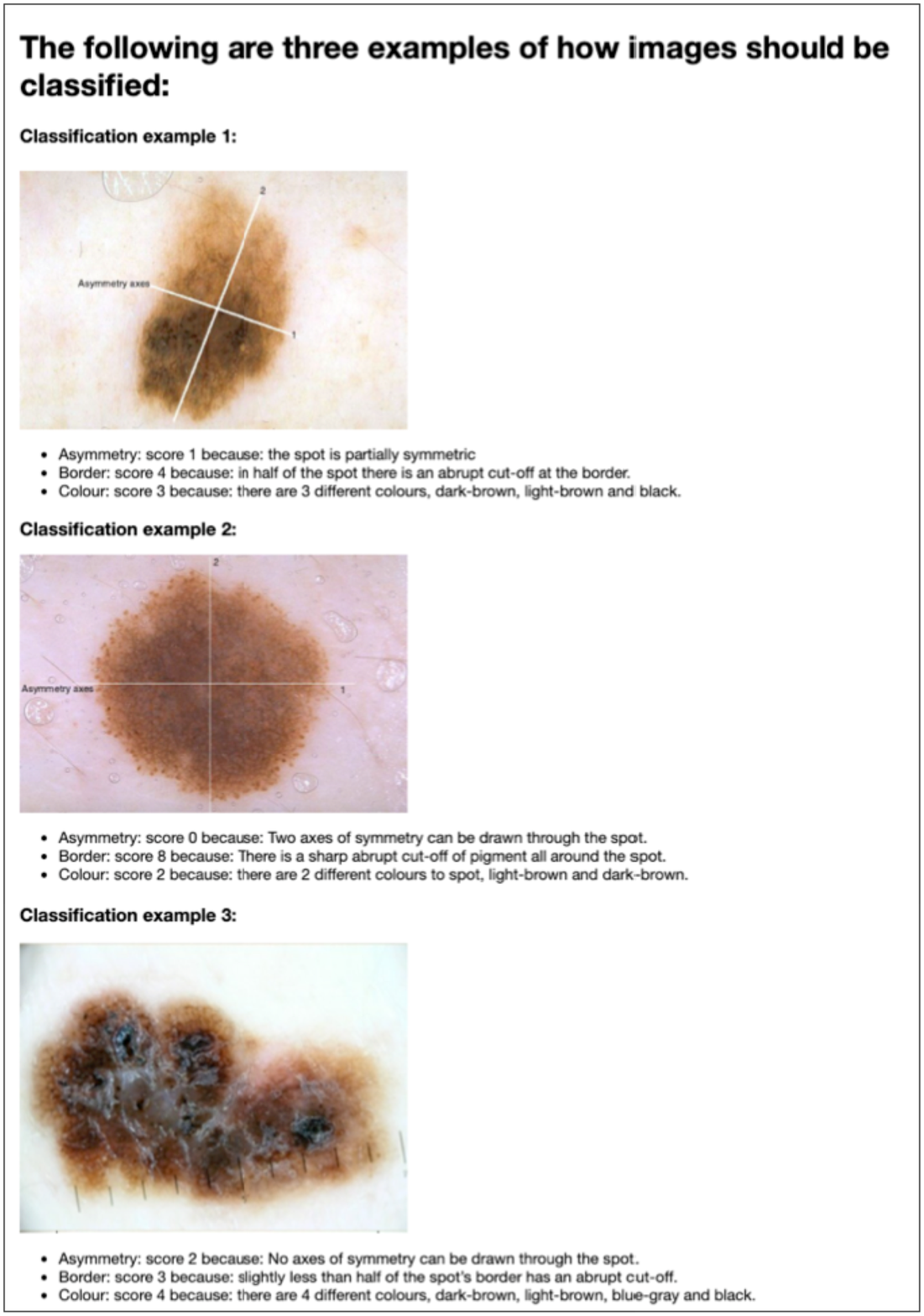}
    \caption{A screen with three examples showing how to score skin lesions.}
    \label{fig:mturk_examples}
\end{figure}

\begin{figure}[ht]
    \centering
    \includegraphics[width=0.6\textwidth]{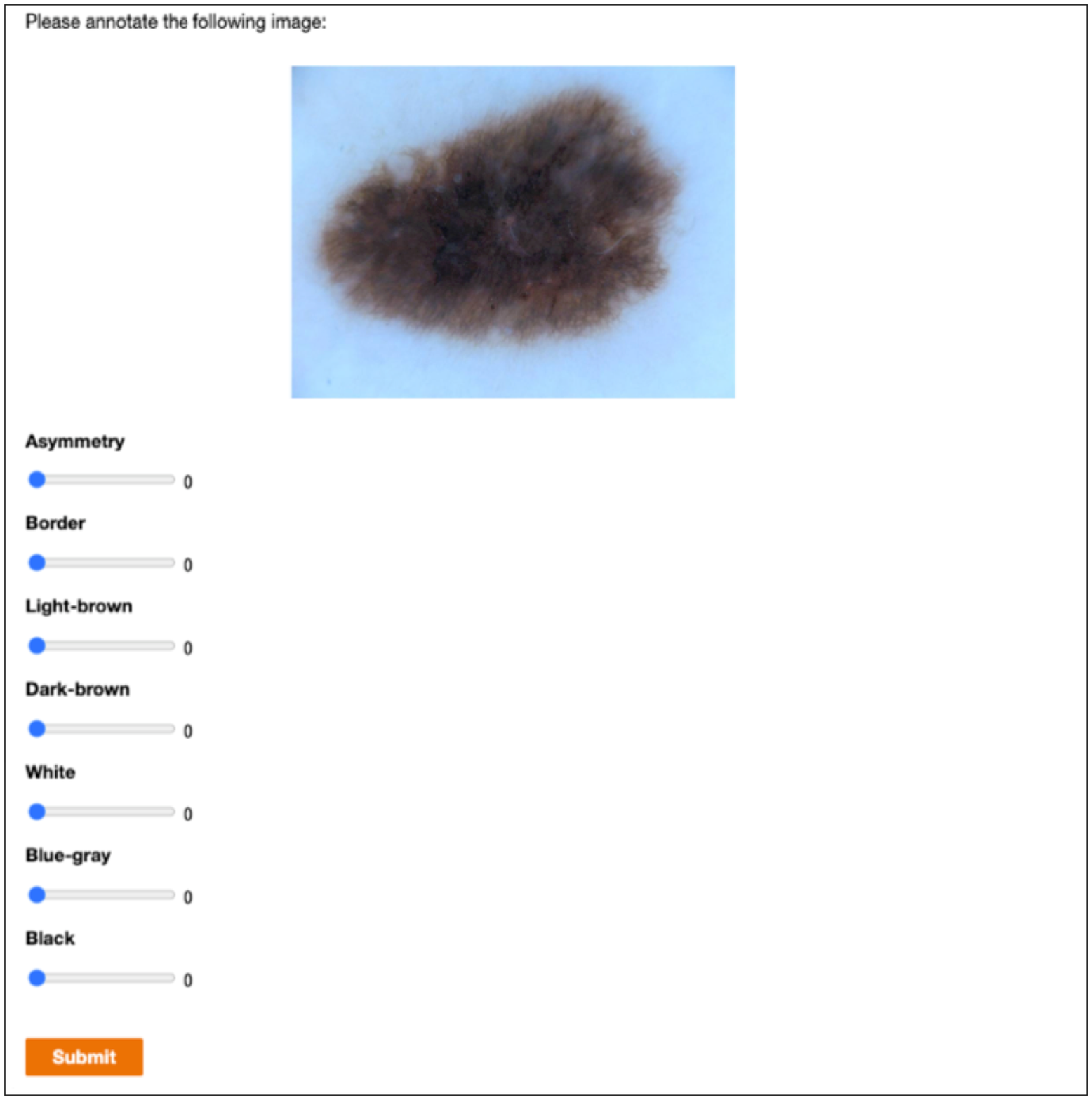}
    \caption{A screen with a skin lesion to be scored using the scoring sliders and pressing the submit button.}
    \label{fig:mturk_annotation}
\end{figure}

The following three figures represent the screens presented to the crowd workers by the online AmazonMechanical Turk (\url{www.mturk.com}) crowdsourcing tool. First, the workers receive instruction (Figure \ref{fig:mturk_instructions}) to assess the skin lesions. Second, the tool presents three examples (Figure \ref{fig:mturk_examples}) that illustrate scoring asymmetry, border and color. Third, the crowdsourcing tool presents one by one the skin lesions (Figure \ref{fig:mturk_annotation}). The crowd workers annotate each presented skin lesion by using the scoring sliders and pressing the submit button. With each skin lesion presented, the MTurk crowdsourcing tool shows the instructions and examples.

\end{document}